\documentclass{article}

\usepackage{microtype}
\usepackage{graphicx}
\usepackage{subfigure}
\usepackage{booktabs} 
\usepackage{lipsum}
\usepackage{tikz}

\usepackage{hyperref}

\usepackage[accepted]{icml2023}

\usepackage{amsmath}
\usepackage{amssymb}
\usepackage{mathtools}
\usepackage{amsthm}

\usepackage[capitalize,noabbrev]{cleveref}

\theoremstyle{plain}

\theoremstyle{definition}

\theoremstyle{remark}

\usepackage[textsize=tiny]{todonotes}

\icmltitlerunning{Unraveling the Complexity of Splitting Sequential Data}

\begin{document}

\twocolumn[
\icmltitle{Unraveling the Complexity of Splitting Sequential Data: Tackling Challenges in Video and Time Series Analysis}



\icmlsetsymbol{equal}{*}

\begin{icmlauthorlist}
\icmlauthor{Diego Botache}{equal,ies}
\icmlauthor{Kristina Dingel}{equal,ies,aim-ed}
\icmlauthor{Rico Huhnstock}{age,aim-ed}
\icmlauthor{Arno Ehresmann}{age,aim-ed}
\icmlauthor{Bernhard Sick}{ies,aim-ed}
\end{icmlauthorlist}

\icmlaffiliation{ies}{Intelligent Embedded Systems, University of Kassel, Wilhelmshöher Allee 73, 34121 Kassel, Germany}
\icmlaffiliation{age}{Institute of Physics and CINSaT, University of Kassel, Heinrich-Plett-Straße 40, 34132 Kassel, Germany}
\icmlaffiliation{aim-ed}{AIM-ED – Joint Lab Helmholtz-Zentrum für Materialien und Energie, Berlin (HZB) and University of Kassel}

\icmlcorrespondingauthor{Diego Botache}{diego.botache@uni-kassel.de}
\icmlcorrespondingauthor{Kristina Dingel}{kristina.dingel@uni-kassel.de}

\icmlkeywords{Machine Learning, ICML, train-test-split, data-centric machine learning, sequential data, videos, time series, splitting, }

\vskip 0.3in
]



\begin{abstract}
\vspace{-.05cm}

Splitting of sequential data, such as videos and time series, is an essential step in various data analysis tasks, including object tracking and anomaly detection. However, splitting sequential data presents a variety of challenges that can impact the accuracy and reliability of subsequent analyses. This concept article examines the challenges associated with splitting sequential data, including data acquisition, data representation, split ratio selection, setting up quality criteria, and choosing suitable selection strategies. We explore these challenges through two real-world examples: motor test benches and particle tracking in liquids.



\end{abstract}

\vspace{-.7cm}

\section{Introduction}
\label{intro}

\vspace{-.2cm}



Sequential data, such as time series and video data, are crucial for technical systems and require accurate modelling of time dependencies to predict future behaviour and identify meaningful patterns. Understanding how variables change over time can gain insights into underlying mechanisms driving the dynamic behaviour of a particular system. Analysis of sequential data can furthermore assist in detecting anomalies or identifying causal relationships. Although one may use basic modelling strategies such as differential equations and stochastic processes, computational limitations arise when dealing with high-dimensional data.
While modelling and analyzing sequential data is crucial in multiple application areas~\cite{barra_deep_2020,martinez_time_2022, masini_machine_2023},
there is currently no default approach concerning splitting sequential information for data-driven algorithms. Moreover, there is a lack of research on evaluating the generalization ability of sequential models trained for specific tasks and assessing the data quality used to train, e.g., machine learning (ML) models. The upcoming section will summarise the main research challenges regarding modelling time dependencies in technical systems, especially during the training phase of an ML cycle. 
We will then frame these challenges in two use cases, namely real-world applications for time series and video data analysis.


\begin{figure}[t!]
    \vspace{-0.1cm}
    \centering
    \includegraphics[width=.85\linewidth]{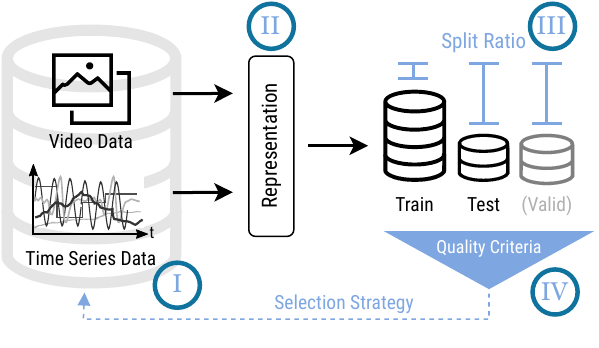}
    \vspace{-0.5cm}
    \caption{We present a path towards a comprehensive data splitting strategy that covers all stages, from data acquisition/selection to finding a suitable representation for machine learning. Our approach involves dividing the data based on an appropriate split ratio and applying quality criteria tailored to a specific use case.}
    \label{fig:abstract}
    \vspace{-0.5cm}
\end{figure}

\vspace{-.2cm}
\section{Splitting Sequential Data: Challenges}
\label{sec:splitting_seq_data}
\vspace{-.2cm}
The common 80\% and 20\% splitting strategy for non-sequential data to train and test ML routines, inspired by the Pareto principle, is unsuitable for modelling time dependencies as it lacks temporal information and dependencies.
One common modification to split sequential data is to use a \textbf{temporal split}, where the first percentage of data, e.g., 80 \%, is used as training data; the remaining data is a test set. This way, one keeps the temporal order of the data. 
This approach, however, may lack generalization of the overall process.
\textbf{Cross-validation}~\cite{Bergmeir2012} can be applied to sequential data, extracting multiple segments for a predefined number $k$ of groups, e.g., folds. This is particularly useful for hyperparameter optimisation considering the average performance. Still, cross-validation methods assume that each sequence in the dataset is independent and identically distributed (i.i.d), which is not valid for correlated time series data. As a result, common cross-validation methods can lead to overly optimistic estimates of the model's performance. 
The \textbf{sliding window approach}~\cite{keogh_online_2001} is another standard method for segmenting sequential data for ML model training. However, critical factors are selecting an optimal window size that captures relevant information and determining the overlap between consecutive windows to avoid missing transitions and patterns.
Given the various options, the question of how to choose the appropriate data selection method for ML training arises. This article states four key aspects towards this goal (cf.~Fig.~\ref{fig:abstract}).
However, it is vital to note that selecting a method depends on the specific task and should be evaluated before addressing the following points. Additionally, we assume that there are no data limitations and that new data can be obtained through simulations or additional experiments. Lastly, we assume that the labeling process has already been completed.

\textbf{I. Data Acquisition / Selection:} Prior to starting the ML process, the experiment's use cases, conditions, restrictions, and boundaries need to be evaluated. The purpose is to establish a sufficient and diverse database containing all relevant information either from experimentation or simulation. This involves considering the system's operational areas and design parameter constraints in the experiment. 

\textbf{II. Representation:} 
To prepare data for ML training (next to preprocessing, of course), one must consider the most fitting representation for the desired experiment. It is essential to evaluate if the parameters of this representation, e.g., sequence length, features, and periodicity, are sufficiently taken into account for training the model and accomplishing the defined task.

\textbf{III. Split Ratio:} 
In general, all splits should contain all relevant scenarios in the experiment to avoid overfitting. High data versatility is necessary to prevent bias during training. Lastly, an isolated validation dataset is recommended when tuning the model's hyperparameters to improve generalization performance.

\textbf{IV. Quality Criteria:} Their definition (in advance) is essential, depending on the use case. Before placing an instance in a particular data split, tools such as statistical tests and distance measures can be used to assess the information gain or similarity of individual pre-selected samples. Addressing changes in the environment, i.e. domain shifts, during experiments and identifying new operation areas are necessary for updating the training data. Task-dependent criteria for change point detection, novelty and anomaly detection are critical for setting up a database which provides enough information about specific scenarios. If it is decided that there are not enough representatives for a scenario, new ones must be sampled/generated considering further selection strategies, cf.~\textbf{I}.

\subsection*{Video Data Use Case: Particle Tracking in Liquids}
Tracking in video data is a common challenge in computer vision tasks where one wants to track an object through a series of images or, in this case, extract the centre coordinates $(x, y, z)$ of particles moving in liquids~\cite{Dingel2021adapt} to implement lab-on-a-chip technologies~\cite{Lim2010}. 
Unlike video data captured in the everyday world, additional factors in microscopic video data can significantly affect tracking, i.e., depth of focus, motion and blur, noise, artefacts, lighting, video binning, and contrast. 
In addition to that, the same issues as with classical object tracking tasks apply, such as missing labels and the difficulty of incorporating simulation data due to the large gap between simulation and reality~\cite{Dingel2021}.
Further, it must be noted that video content can vary significantly from experiment to experiment. But also, consecutive video frames within one experiment may appear similar due to high frame rates or long sequences without particle movement. It leads to imbalanced datasets, and selecting relevant video sequences for training and testing is of utmost importance~\textbf{(I,~IV)}. Especially in the case of tracking, attention must be paid to the sequence length~\textbf{(II)} considering the possible shortage of storage space during the recording of an experiment, which results in short video sequences whose separation would not be ideal. This makes it all the more important to consider an appropriate data representation~\textbf{(II)} and split ratio~\textbf{(III)} and finally establish data quality criteria, which in turn will largely dictate the selection strategy for extending the data splits.
Finally, this particular use case shows the benefit of having a closed-loop-experimentation station~\cite{Sanchez-Lengeling2018}. It allows the creation of new training data on the fly by actively~\cite{Kottke2021} asking for missing or underrepresented data using the pre-defined quality criteria and requirements~\textbf{(IV)}. 
 

\subsection*{Time Series Data Use Case: Motor Test Bench}
This use case involves monitoring a motor test bench using deep-learning techniques~\cite{Botache2021}. The data is high-dimensional and heterogeneous, obtained from multiple sensors capturing factors affecting motor performance. An ML-supported monitoring strategy must consider correlations and time dependencies to detect faults early on, improving the experimental process. 
When designing fault detection as a binary classification task, the data should contain faulty instances for each operating area. Also, acquiring multiple measurements independently from each other is critical to generate splits containing single experiments. Therefore, the i.i.d. assumption should hold for each data split~\textbf{(I)}. Extracting segments of the same length is the first step for creating the input space for ML techniques~\textbf{(II)}. Still, long sequences represent higher model complexity for a constant sampling rate and shorter sequences, on the other side, may not capture enough time dependencies in the data. Next, extracting significant features could reduce the complexity of the task, e.g., by removing categorical signals and employing representation learning techniques. 
In a binary classification task, fault instances are usually rare, challenging to acquire, and therefore underrepresented in data splits. Oversampling techniques~\cite{kovacs_empirical_2019}, active-learning with class-balance selection~\cite{cai2022active}, and synthetic data generation are highly recommended for improving the split ratio of the datasets~\cite{Westmeier2022}~\textbf{(III)}. In addition, using weighted losses~\cite{lin2017focal} can improve the learning task of the algorithms. Finally, the quality assessment of the data split using similarity measurements is necessary for identifying potentially unexploited areas in the data distribution~\textbf{(IV)}.

\section{Conclusion}

Especially with respect to ML training processes, extracting appropriate training and test splits for time-dependent data is still a major challenge. In this concept paper, we identified four key research questions and put them in concrete terms using two use cases for time series and video data, which now require further investigation.

\section*{Acknowledgements}

We gratefully acknowledge the assistance and support of the Joint Laboratory Artificial Intelligence Methods for Experiment Design (AIM-ED) between Helmholtzzentrum für Materialien
und Energie, Berlin, and the University of Kassel.



\bibliography{bib/bibliography}
\bibliographystyle{icml2023}



\end{document}